\documentclass[11pt]{article}

\usepackage[margin=1in]{geometry}
\usepackage[T1]{fontenc}
\usepackage[utf8]{inputenc}
\usepackage{lmodern}
\usepackage{microtype}
\usepackage{amsmath,amssymb}
\usepackage{booktabs}
\usepackage{multirow}
\usepackage{graphicx}
\usepackage[numbers,sort&compress]{natbib}
\usepackage[colorlinks=true,linkcolor=blue,citecolor=blue,urlcolor=blue]{hyperref}
\usepackage{xcolor}
\hypersetup{
  pdftitle={Training-Free Off-Screen Player Imputation for Broadcast-Based Spatial Football Analytics},
  pdfauthor={Seongjin Choi},
  pdfsubject={Off-screen player imputation and spatial football analytics},
  pdfkeywords={football analytics, player imputation, pitch control, game state reconstruction}
}

\newcommand{\pp}{\,pp}   

\title{Training-Free Off-Screen Player Imputation for\\Broadcast-Based Spatial Football Analytics}

\author{%
  Seongjin Choi\thanks{Independent researcher.
  ORCID: \url{https://orcid.org/0009-0001-3193-7424}.
  Code and data: \url{https://github.com/nowayfootball/offscreen-impute}.
  Contact: \texttt{nowayfootball@gmail.com}.}%
}
\date{}

\begin{document}
\maketitle

\begin{abstract}
Spatial football metrics such as pitch control assume access to the positions of
all 22 players, yet the most widely available source of positional data --- the broadcast main
camera --- shows only 10--16 of them at any moment. We quantify the resulting
distortion with an open, reproducible benchmark: a simulated broadcast viewport
applied to open full-pitch tracking data (Metrica Sports; three matches, one
held out from method development). Ignoring off-screen
players --- the visible-only baseline implied whenever a video-based
game-state-reconstruction (GSR) pipeline adds no imputation layer --- inflates
hidden-zone pitch-control error to 25.1--26.9 percentage points and produces a
mean absolute control-share error of 11.1--13.4 points across the three
matches. We then evaluate a ladder of
\emph{training-free, online} imputation baselines that use only observations from
the match being analysed. The best overall on these decision-relevant metrics, \emph{role-anchored centroid
voting} --- each visible player votes for the full-team centroid by subtracting
its running role offset, attenuating the viewport-induced subset bias ---
roughly halves hidden-zone error (to 12.2--13.8 points) and cuts the
control-share error to 28--48\% of the ignore policy at every viewport width
from 36\,m to 60\,m in all three matches (4.5--4.7 points at $W=44$\,m). In
the short-occlusion regime ($\le$9.6\,s) covered by the closest learned prior
work, our training-free method reaches binwise median position errors of
3.3--8.9\,m; 50--57\% of hidden-player observations under the simulated
viewport, however, lie beyond that regime, outside the 9.6\,s sequence
protocol of that prior work. We integrate the
method end-to-end into a broadcast-video GSR pipeline and show that
imputation changes a downstream possession-quality score (Space-Creation
Index) by 15.6 and 17.2 points on two real World Cup broadcast windows,
flipping the verdict class --- under prespecified operational thresholds ---
in one of them.
\end{abstract}

\section{Introduction}

Video-based game state reconstruction (GSR) turns ordinary broadcast footage into
player coordinates on a common pitch template, promising tracking-level analytics
without stadium hardware~\citep{somers2024soccernet}. But the broadcast main
camera pans and zooms with the ball: in our World Cup broadcast clips only
10--16 of the 22 players are visible in a typical frame, consistent with the
9--12 reported for broadcast video by~\citet{ochin2025footpass} and the
12.8$\pm$3.7 of~\citet{omidshafiei2022}.
Team-level spatial metrics --- pitch control~\citep{spearman2017physics,
spearman2018beyond}, space value, block compactness --- are then computed over a
biased subset of players: the team whose defenders sit off-screen silently cedes
control of the hidden half of the pitch, not because of anything that happened on
the pitch, but because of where the camera pointed.

This distortion is routinely acknowledged and rarely measured. The GSR evaluation
protocol itself (GS-HOTA~\citep{somers2024soccernet}) scores only the players
that are visible; what happens to downstream \emph{team-level} metrics when the
invisible remainder is ignored has, to our knowledge, not been quantified on open
data. The one closely related study, DeepMind's Graph
Imputer~\citep{omidshafiei2022}, demonstrated the problem and a learned solution,
but on 105 proprietary Premier League tracking matches, with a bidirectional
model that consumes \emph{future} observations inside 9.6-second windows ---
neither reproducible nor applicable in a strictly online setting.

This paper makes three contributions:

\begin{enumerate}
\item \textbf{A reproducible distortion benchmark.} We simulate a broadcast
viewport (ball-tracking pan with lag, width $W$) over open full-pitch tracking
data and score any imputation policy by hidden-zone pitch-control MAE, team
control-share error, and hidden-player position error --- metrics defined by the
downstream analytics rather than by trajectory fidelity alone
(Section~\ref{sec:benchmark}).
\item \textbf{A ladder of training-free online baselines}, culminating in
\emph{role-anchored centroid voting} (B4), which halves the hidden-zone control
error and cuts control-share error to 28--48\% of the ignore-policy
baseline, without offline training, future observations, or GPU
(Sections~\ref{sec:ladder}--\ref{sec:results}).
\item \textbf{End-to-end integration} into a video$\to$GSR$\to$metrics
pipeline, with a case study on real World Cup broadcasts where imputation
moves a possession-quality score by a decision-relevant magnitude, changing
the verdict class in one of two windows (Section~\ref{sec:application}).
\end{enumerate}

Our position is deliberately modest: learned imputers remain preferable when
training data and future context are available. What we establish is (i) how
large the distortion actually is under a simulated viewport, (ii) a
training-free floor that learned methods should beat, and (iii) an explicit
scoring of the long-occlusion regime ($>$9.6\,s) that lies outside the
sequence protocol of the closest prior work --- and turns out to hold roughly
half or more (50--57\%) of hidden-player observations in our benchmark.

\section{Related Work}
\label{sec:related}

\paragraph{Learned trajectory imputation.}
The Graph Imputer of \citet{omidshafiei2022} is the closest prior work: a
graph-network variational model trained on 105 proprietary EPL tracking matches,
evaluated with a simulated camera view-cone, and applied to pitch control under
partial observability --- the same problem framing as ours. Two protocol
differences preclude direct numeric comparison, and we analyse them explicitly in
Section~\ref{sec:protocol}: their model is bidirectional (it consumes future
observations within 9.6\,s windows), and trajectories are segmented into
9.6\,s sequences --- bounding the occlusion span the model must bridge ---
whereas our setting is strictly online with unbounded occlusions.
MIDAS~\citep{choi2025midas} is a recent learned set-transformer imputer
reporting strong results on multi-agent trajectory imputation, including
camera-mask experiments; TranSPORTmer~\citep{capellera2024transportmer} similarly
unifies forecasting, imputation, and agent-status inference for multi-agent
sports in a single transformer under an input mask. \citet{everett2023inferring}
impute full-team player locations from sparse event observations using recurrent
and graph-based components, enabling downstream analyses such as pitch dominance;
their observation setting is event-driven rather than camera-mask-driven and
requires offline training, whereas our benchmark targets causal broadcast-view
occlusion without training data. \citet{penn2025continuous} reconstruct continuous
trajectories from discrete broadcast-derived observations with a parsimonious
autoregressive (ARMAX) model that uses the ball position as an exogenous
input --- an adjacent observation model where sparsity is temporal rather than
spatial.

\paragraph{Ghosting.}
The ghosting line of work~\citep{le2017ghosting, yurko2024nflghosts} generates
counterfactual league-average positions --- where a player \emph{should} have
been --- for coaching evaluation. The goal differs from ours: ghosting evaluates
decisions against a baseline policy, whereas imputation recovers actual
unobserved positions for measurement.

\paragraph{Game state reconstruction.}
SoccerNet-GSR~\citep{somers2024soccernet} defines the task and the GS-HOTA
metric for recovering \emph{visible} players from broadcast video; camera
calibration~\citep{gutierrez2024pnlcalib, falaleev2024keypoints} and multi-object
tracking~\citep{aharon2022botsort} supply the geometry and identity backbone.
FOOTPASS~\citep{ochin2025footpass} reports 9--12 visible players per broadcast
frame and notes imputation as a pipeline necessity, corroborating the
universality of the problem. Our benchmark scores the invisible remainder that
GS-HOTA leaves undefined.

\section{Benchmark}
\label{sec:benchmark}

\paragraph{Data.}
Metrica Sports open tracking~\citep{metrica2020}: three matches, all 22
on-pitch players plus ball at 25\,fps (games 1--2 ship as full-match CSVs;
game 3, in the EPTS-FIFA format, covers one half). We use the first 45 minutes
of each and evaluate at 5\,fps; individual player samples with missing
coordinates are dropped (frames are kept). Games 1--2 were used for method and
hyperparameter development; game 3 --- and the ablation variants B3E/B3V/B5
--- were specified and frozen before game 3 was first evaluated, so game 3
serves as a held-out check of the method ladder (the interpretive text was, of
course, written after seeing all results). Using full-pitch tracking as ground truth lets us compute
the \emph{true} spatial metrics and score any partial-observation policy
against them.

\paragraph{Viewport simulator.}
A virtual main camera pans horizontally following an exponentially smoothed ball
position ($\alpha = 0.06$ per frame at 25\,fps), mimicking broadcast pan lag; the visible region is a width-$W$ window spanning the full
pitch height. At $W = 44$\,m the viewport shows 14.6--15.0 players on average
(games 1--3), within the range of our real broadcast clips (10--16 visible) and
close to the 12.8$\pm$3.7 reported by~\citet{omidshafiei2022}. Sensitivity is reported for
$W \in \{36, 44, 52, 60\}$\,m.

\paragraph{Metrics.}
We score three quantities, chosen to reflect the downstream analytics rather
than trajectory fidelity alone:
\begin{enumerate}
\item \textbf{Hidden-player position error} (m): distance between imputed and
true position, over all (frame, player) pairs where the player is outside the
viewport and has been observed at least once earlier in the half
(never-observed players cannot be scored for position; they remain absent from
the control maps of every partial-observation condition --- the ground-truth
map always contains them --- so the map metrics do include the cold-start
penalty). Occlusion gap denotes elapsed time since the player's last
observation at each hidden frame; occlusion-share denominators use this same
population.
\item \textbf{Pitch-control map MAE}: pitch control is computed on a 3\,m grid
with an arrival-time model and sigmoid contest
function~\citep{spearman2017physics}; we report mean absolute error against the
full-observation control map, over the full pitch and over the \emph{hidden
zone} (grid cells outside the viewport) separately. Control is computed with
zero velocities in all conditions, isolating the effect of imputation from that
of velocity estimation; the reported magnitudes therefore apply to this
position-only control variant (velocity-aware control is discussed in
Section~\ref{sec:limitations}).
\item \textbf{Team control-share error} (percentage points): mean absolute
per-frame deviation of the single headline number ``team A controls $x\%$ of
the pitch''.
\end{enumerate}
We designate (1) and (3) as the \emph{decision-relevant} metrics --- they
determine where players are placed and what share is reported --- and select
among methods on them; the map MAE (2) is reported for completeness. The hidden-zone MAE and the share error carry 95\%
block-bootstrap confidence intervals over one-minute blocks (frames within a
block are strongly autocorrelated, so frame-level resampling would be
anticonservative); these within-match intervals quantify temporal sampling
uncertainty, not match-to-match variation.

\section{The Imputation Ladder}
\label{sec:ladder}

All methods below are \emph{causal} (no future observations), use only
observations from the current match (no offline training), and run faster than
real time on a single CPU core (closed-form arithmetic, no learned components). Let $V_t$ denote the set of visible players at time $t$, $p_i(t)$ the
position of player $i$, and $\mathrm{off}_i(t)$ a running estimate of player
$i$'s \emph{role offset} --- its displacement from the team centroid. All
quantities are maintained separately per team (including the three-voter
threshold below); we drop the team index for readability.

\begin{description}
\item[B0 --- ignore.] Hidden players are absent from the metric --- the
visible-only baseline implied whenever no imputation layer is added; GSR
evaluation itself covers visible players only~\citep{somers2024soccernet}.
\item[B1 --- last-seen with decay.] $\hat{p}_j = w\,p_j^{\mathrm{last}} +
(1-w)\,\bar{p}_V$ with $w = e^{-\Delta t/\tau}$, $\tau = 8$\,s: hold the last
observed position, decaying toward the visible-team mean.
\item[B2 --- formation anchor.] While player $i$ is visible, store
$\mathrm{off}_i = p_i - \bar{p}_{V}$, its offset from the \emph{visible-team}
centroid (a visible-subset offset --- the viewport bias it inherits is what B4
removes). When hidden, place the player at the current visible centroid plus
the stored offset.
\item[B5 --- fixed formation template.] As B2, but the offset is the
cumulative mean over all of player $i$'s visible frames so far, rather than
the latest snapshot --- the closest online analogue of a static role template.
\item[B3 --- B2 + EMA offsets + velocity extrapolation.] As B2, but with
exponentially averaged offsets --- stored, like B4's, relative to the voted
centroid of Eq.~(1) when available, yet \emph{applied} at the plain visible
centroid --- and constant-velocity extrapolation for recently hidden players,
blended into the anchor with weight $e^{-\Delta t/1.5\,\mathrm{s}}$.
Its two ingredients are isolated as B3E (EMA only) and B3V (velocity only) in
the ablation (Section~\ref{sec:results}).
\item[B4 --- role-anchored centroid voting.] The visible-team centroid
$\bar{p}_{V}$ in B2 is itself biased by the viewport: when the camera points
left, the visible subset sits left of the true team centroid, and every stored
offset inherits that bias. B4 attenuates it by \emph{voting}: each visible player
proposes the full-team centroid as its position minus its role offset,
\begin{equation}
\hat{c}(t) \;=\; \frac{1}{|V_t|}\sum_{i \in V_t}\bigl(p_i(t) - \mathrm{off}_i(t^-)\bigr),
\qquad
\mathrm{off}_i(t) \;\leftarrow\; \mathrm{EMA}\bigl[\,p_i(t) - \hat{c}(t)\,\bigr]
\quad (i \in V_t),
\end{equation}
where $\mathrm{off}_i(t^-)$ denotes the offset value \emph{before} the update
at $t$: each step votes with the previous offsets, then updates them against
the voted centroid (EMA weight $0.1$ per 5\,fps evaluation step). This is a self-consistent bootstrap --- offsets are stored
relative to the voted centroid, and the voted centroid is computed from those
same offsets. The recursion is seeded from the visible mean until three
offset-bearing players are available; thereafter, newly appearing players are
initialised relative to the voted centroid. Hidden player $j$ is imputed at
$\hat{c}(t) + \mathrm{off}_j$. Because each player's individual viewport bias
is subtracted before averaging, the systematic subset bias is attenuated rather
than accumulated (exact cancellation would require stable offsets and
conditionally representative voters). When fewer than three voters are available, B4 falls back to the
formation anchor (B2), and B2 in turn to last-seen (B1) for players without a
stored offset; such fallback frames are included in the reported numbers. At
$W = 44$\,m the voting path produces 70--86\% of B4's hidden-player estimates
(B2 fallback 14--30\%, B1 0\%), so the reported performance is a mix in those
proportions.
\end{description}

\section{Results}
\label{sec:results}

\begin{table}[t]
\centering
\caption{Imputation ladder at $W = 44$\,m (first halves of the three Metrica
matches; game 3 held out from method development). Lower is better
throughout; bold marks the best method per game and metric. B0 has no position
error since it places no players.}
\label{tab:ladder}
\small
\begin{tabular}{lcccc}
\toprule
& \shortstack{hidden ctrl MAE (pp)\\g1 / g2 / g3} & \shortstack{full-pitch MAE (pp)\\g1 / g2 / g3} & \shortstack{share err.\ (pp)\\g1 / g2 / g3} & \shortstack{median pos.\ err.\ (m)\\g1 / g2 / g3} \\
\midrule
B0 ignore                    & 26.9 / 25.6 / 25.1 & 18.6 / 17.2 / 16.7 & 13.4 / 12.5 / 11.1 & --- \\
B1 last-seen                 & 22.1 / 20.0 / 19.5 & 16.2 / 14.2 / 13.5 & 10.6 / 9.5 / 8.2   & 19.6 / 17.9 / 18.4 \\
B2 anchor                    & 15.7 / 14.6 / 14.3 & 12.2 / 10.8 / 10.3 & 6.2 / 5.4 / \textbf{4.4} & 13.6 / 12.8 / 12.5 \\
B5 fixed template            & 23.0 / 18.8 / 19.1 & 17.6 / 14.2 / 14.2 & 10.3 / 7.7 / 6.9   & 22.7 / 20.7 / 21.9 \\
B3E ablation: EMA only       & 13.2 / 12.2 / 13.6 & 10.8 / 9.8 / 10.5 & 5.8 / 4.8 / 5.0    & 15.7 / 15.2 / 16.3 \\
B3V ablation: velocity only  & 15.5 / 14.4 / 14.0 & 12.0 / 10.6 / 10.0 & 6.2 / 5.5 / \textbf{4.4} & 13.2 / 12.2 / 11.9 \\
B3 EMA + velocity            & \textbf{12.8} / \textbf{11.8} / \textbf{13.2} & \textbf{10.4} / 9.3 / 10.0 & 5.7 / 4.6 / 4.7 & 14.6 / 14.0 / 15.1 \\
B4 centroid vote             & 13.3 / 12.2 / 13.8 & 10.5 / \textbf{9.2} / \textbf{9.9} & \textbf{4.7} / \textbf{4.5} / 4.7 & \textbf{11.6} / \textbf{10.0} / \textbf{9.7} \\
\bottomrule
\end{tabular}
\end{table}

\begin{figure}[t]
\centering
\includegraphics[width=\linewidth]{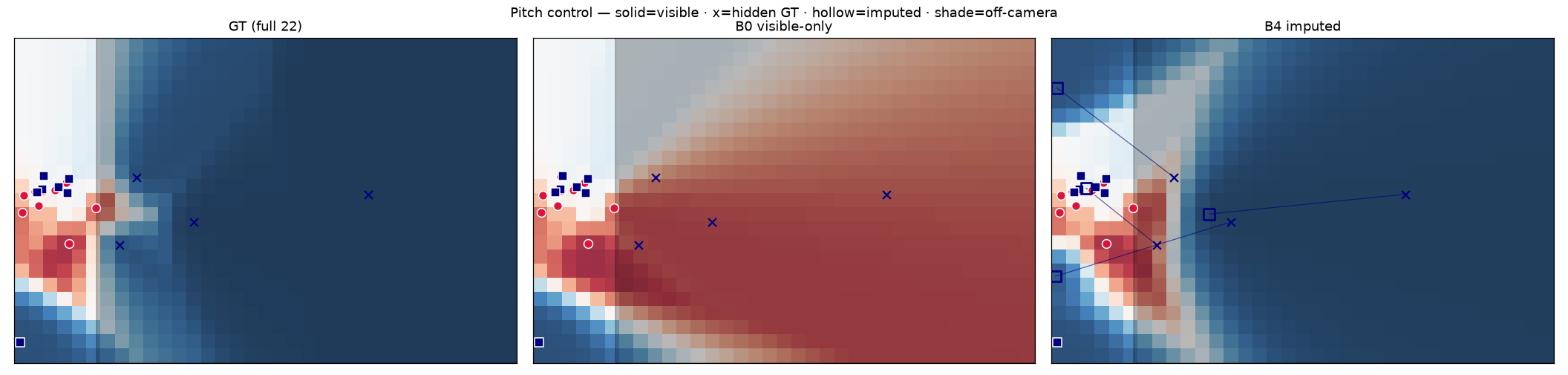}
\caption{Benchmark visualisation (Metrica, $W = 44$\,m). Left: ground-truth
pitch control from all 22 players. Middle: control from visible players only
(B0) --- the hidden zone (outside the vertical viewport band) is systematically
misassigned. Right: control with B4 imputation (ghost markers) --- the hidden
zone is largely restored without any training data.}
\label{fig:panels}
\end{figure}

\paragraph{Ladder.}
Table~\ref{tab:ladder} reports the ladder at $W = 44$\,m.
Ignoring off-screen players (B0) leaves a hidden-zone control MAE of
25.1--26.9\pp{} and leaves a mean absolute control-share error of 11.1--13.4\pp{}. The naive last-seen policy removes only about a fifth to a quarter of
it. The formation anchor (B2) is the single largest step --- most of a hidden
player's position is explained by ``the team moved, the player kept its
role''. Centroid voting (B4) then attenuates the viewport subset bias that B2
inherits. It is the best method on position error in all three matches
(9.7--11.6\,m median vs.\ 12.5--13.6\,m for B2) and on share error in games
1--2 (4.5--4.7\pp{}); in the held-out game 3, B2 and B3V edge it on share
error by 0.3\pp{} (4.4 vs.\ 4.7) --- a paired one-minute block-bootstrap
95\% interval for the game-3 B4$-$B2 share-error difference is
$[-0.4, +1.3]$\pp{} and includes zero, so this gap is not resolved at this
sample size. Full-pitch MAE
(Table~\ref{tab:ladder}) improves in step. Marginal block-bootstrap 95\%
intervals for B0 vs.\ B4 do not overlap in any match on hidden-zone MAE or
share error (e.g., game 1 hidden-zone MAE: B0 $[24.5, 29.4]$ vs.\ B4
$[11.2, 15.6]$; share error $[11.4, 15.7]$ vs.\ $[3.6, 6.1]$). Figure~\ref{fig:panels} visualises the effect on a single
frame: the hidden zone misassigned by B0 is largely restored by B4.

\paragraph{Ablation: what actually helps.}
The B3E/B3V decomposition settles what drives B3 --- and corrects a reading we
ourselves held from the confounded B2-vs.-B3 comparison. Velocity blending
alone (B3V) \emph{slightly improves} position error over B2 in all three
matches (13.6$\to$13.2, 12.8$\to$12.2, 12.5$\to$11.9\,m median): short-gap
extrapolation is mildly helpful, not harmful. EMA offsets applied at the plain
visible centroid (B3E) give the second-lowest hidden-zone map MAE in all three
matches (B3 is lowest) but the worst position error of the dynamic-anchor
variants (15.2--16.3\,m): the offsets are accumulated against the voted
reference of Eq.~(1) yet applied at the biased visible centroid, and the
mismatch misplaces individual players. B4 applies the same EMA offsets at the
voted centroid instead, making estimation and application self-consistent, and
obtains the best position error. B3E and B3 are best read as diagnostic
reference-mismatch variants --- both estimate offsets against the voted centroid
but apply them at the visible centroid --- rather than as natural competing
baselines. The clearest negative result is the fixed
template (B5), far worse than any dynamic anchor (position error
20.7--22.7\,m; share error 6.9--10.3\pp{}) --- consistent with within-half
role drift, though slow adaptation and accumulated subset bias may also
contribute.

\begin{figure}[t]
\centering
\includegraphics[width=\linewidth]{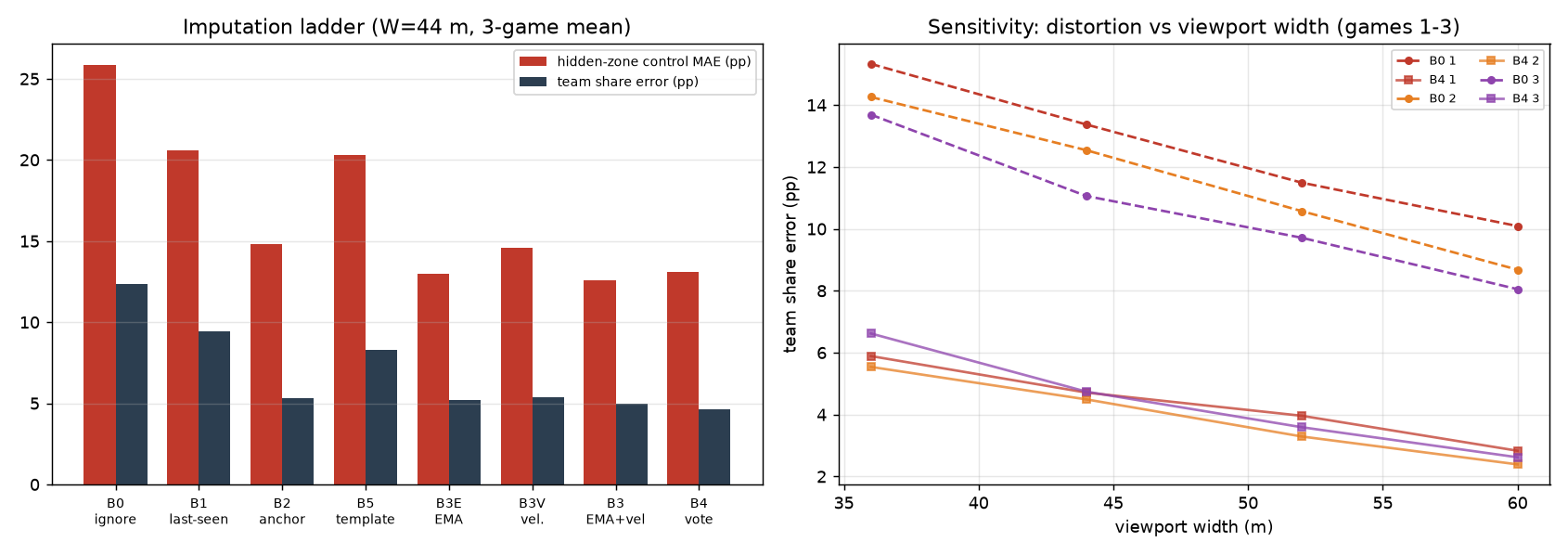}
\caption{Left: imputation ladder at $W=44$\,m --- three-game mean of
hidden-zone control MAE and control-share error, all eight methods. Right:
viewport-width sensitivity of the share error --- B4 (solid) vs.\ B0 (dashed)
across $W = 36$--$60$\,m, all three games. B4's absolute share-error gain is
largest in the narrowest (hardest) viewports.}
\label{fig:ladder}
\end{figure}

\begin{table}[t]
\centering
\caption{Viewport-width sensitivity (game 1 / game 2 / game 3): hidden-zone
control MAE and control-share error, B0 vs.\ B4.}
\label{tab:sens}
\small
\begin{tabular}{cccccc}
\toprule
& & \multicolumn{2}{c}{B0 ignore} & \multicolumn{2}{c}{B4 centroid vote} \\
\cmidrule(lr){3-4}\cmidrule(lr){5-6}
$W$ (m) & visible (of 22) & MAE (pp) & err.\ (pp) & MAE (pp) & err.\ (pp) \\
\midrule
36 & 13.0 / 13.1 / 13.1 & 28.9 / 27.6 / 28.3 & 15.3 / 14.3 / 13.7 & 15.6 / 14.1 / 16.8 & 5.9 / 5.5 / 6.6 \\
44 & 14.6 / 14.9 / 15.0 & 26.9 / 25.6 / 25.1 & 13.4 / 12.5 / 11.1 & 13.3 / 12.2 / 13.8 & 4.7 / 4.5 / 4.7 \\
52 & 16.1 / 16.3 / 16.5 & 24.4 / 23.0 / 23.3 & 11.5 / 10.6 / 9.7 & 11.6 / 9.8 / 12.1 & 4.0 / 3.3 / 3.6 \\
60 & 17.2 / 17.4 / 17.7 & 22.8 / 20.6 / 21.2 & 10.1 / 8.7 / 8.1 & 9.5 / 8.0 / 10.5 & 2.8 / 2.4 / 2.6 \\
\bottomrule
\end{tabular}
\end{table}

\paragraph{Sensitivity.}
Table~\ref{tab:sens} and Figure~\ref{fig:ladder} report the sweep over viewport
widths. Across $W = 36$--$60$\,m B4 improves on B0 in every cell of all three
matches, reducing the share error to 28--48\% of B0 across games and widths,
with the largest absolute share-error gains in the
narrowest (hardest) viewports --- the method is most valuable exactly when
observation is poorest.

\subsection{Occlusion-time stratification and protocol comparison}
\label{sec:protocol}

\begin{table}[t]
\centering
\caption{Observation-protocol comparison with the Graph
Imputer~\citep{omidshafiei2022}. The regimes differ in the two dimensions that
matter most --- temporal information and training data --- so numeric results
are not directly comparable; the stratified analysis below partially narrows the
occlusion-regime mismatch.}
\label{tab:protocol}
\small
\begin{tabular}{lll}
\toprule
& Graph Imputer & This work \\
\midrule
Camera model & view-cone $\times$ pitch polygon, ball-led & ball-EMA pan window (width $W$) \\
Visible players & 12.8 $\pm$ 3.7 of 22 & 14.6--15.0 of 22 ($W=44$\,m) \\
Temporal structure & bidirectional (sees \emph{future} obs.), & online, causal, \\
& sequences segmented at 9.6\,s & occlusions unbounded \\
Training data & 105 proprietary EPL matches & none (current match only) \\
Compute & GPU training & CPU, real time \\
\bottomrule
\end{tabular}
\end{table}

Table~\ref{tab:protocol} contrasts the observation protocols, whose
differences are likely favorable to the learned model --- bidirectional
inference over future observations, 9.6\,s sequence segmentation, and 105
matches of training data --- so we align the comparison by stratifying B4's
position error by occlusion duration:

\begin{center}
\small
\begin{tabular}{lccc}
\toprule
occlusion gap & $\le$2\,s & 2--9.6\,s & $>$9.6\,s \\
\midrule
B4 median position error & 3.3--3.7\,m & 7.2--8.9\,m & 15.6--16.9\,m \\
share of hidden samples (frame-weighted) & \multicolumn{2}{c}{$\le$9.6\,s: 43--50\%} & $>$9.6\,s: 50--57\% \\
\bottomrule
\end{tabular}
\end{center}

For occlusion gaps short enough to fit within a 9.6\,s window --- the regime
closest to the Graph Imputer's protocol --- a training-free online method
reaches 3.3--8.9\,m median error. We do not attempt a numeric comparison:
masking, aggregation, and data all differ, and we do not claim superiority over
learned bidirectional models within their own evaluation regime; the point is that a zero-training
method is usable there at all.
Roughly half or more of the hidden (frame, player) samples in our benchmark
(50--57\%, frame-weighted --- long episodes dominate this count) lie
\emph{beyond} 9.6\,s: long occlusions of far-side defenders
during sustained attacking phases are precisely where spatial metrics need
imputation most, and it lies outside the sequence protocol of the closest
prior work. It is also where our absolute errors (15.6--16.9\,m median) show
the remaining headroom, which we expect learned long-horizon models to attack.

\section{Application: Possession-Quality Verdicts on Real Broadcasts}
\label{sec:application}

To probe the effect on real footage, we integrate a B4-inspired ghosting layer
into a broadcast-video GSR pipeline assembled from open-source components: PnLCalib-based camera
calibration~\citep{gutierrez2024pnlcalib} with a temporal bridge, BoT-SORT
tracking~\citep{aharon2022botsort}, colour-based team assignment, and shot-change
gating; the pipeline scores GS-HOTA 35.2 on the 11 publicly downloadable
sequences of the SoccerNet-GSR test split (for context only: the official
baseline reports 22.26 on the full test set~\citep{somers2024soccernet}, so the
two numbers are not strictly comparable; challenge-winning systems with GPU
appearance re-identification score substantially higher, and our pipeline is
CPU-only). Off-screen players
are rendered as ``ghosts'' with role-anchored positions that fade over a 20\,s
lifetime; ghosts top the team up to at most eleven players (visible plus
ghosts), and a ghost is retired when a newly appearing track (within 2\,s of its
birth) emerges within 3\,m of the ghost's prediction. Role offsets are keyed to track identities, so a fragmented track
restarts its offset estimate --- the operational layer is B4-inspired rather
than the benchmarked algorithm verbatim.

As a downstream consumer we use a possession-quality metric, the
\emph{Space-Creation Index} (SCI): during a flagged low-threat possession
(``junk possession''), $\mathrm{SCI} = \Delta_{\mathrm{own}} +
\Delta_{\mathrm{opp}}$, where $\Delta_{\mathrm{own}}$ is the change in the
possessing team's pitch-control share of the attacking third between the first
and last thirds of the window, and $\Delta_{\mathrm{opp}}$ is the collapse of
the opponent's control share in its own advanced zone (both in percentage
points, from the same pitch-control model as
Section~\ref{sec:benchmark}). Operational verdict classes: \emph{space
creation} (SCI $\ge +12$) --- the possession creates space even without
shots; \emph{weak progression} ($+4 \le$ SCI $< +12$); \emph{dead possession}
(SCI $< +4$). We evaluate the two junk-possession windows that our event-side index flags
for a FIFA World Cup 2026 Round-of-32 match between the Netherlands and Morocco
(1--1 after extra time; Morocco won 3--2 on penalties) --- flagged from event
data alone (possession sequences of negligible
threat under an expected-threat grid), before any spatial analysis, and they
are the only flagged windows for this match, so there is no post hoc window
selection. Morocco is the possessing team in both. Window 1 spans game clock
11:58--12:32 (34\,s), window 2 spans 72:48--73:45 (57\,s); scorelines below
are at each window's start. The attacking third and the opponent's advanced
zone are the corresponding thirds of pitch length in each team's attack
direction; shares are frame-averaged over the first and last temporal thirds
of the window, and the SCI class thresholds ($+12$/$+4$) are prespecified
operational bins, set in our earlier internal work on this index before these
windows were analysed but not externally validated. Imputation moves the score
by a decision-relevant magnitude:

\begin{itemize}
\item Window 1 (0--0, min.\ 12): SCI $+15.6 \to +32.8$. Under the imputation
policy, the estimated off-screen defenders produce a substantially
\emph{stronger} space-creation signal than the visible-only estimate.
\item Window 2 (0--1, min.\ 73): SCI $-10.9 \to +4.7$ --- the verdict class
changes from ``dead junk'' to ``weak progression''.
\end{itemize}

Absent full-pitch ground truth for broadcast footage we make no claim that the
imputed scores are more accurate; the point is a sensitivity result. Visible-only spatial
verdicts can move by 15.6--17.2 SCI points under imputation --- more than
spanning the 8-point intermediate verdict class --- so spatial
possession-quality reported from broadcast video without imputation is
sensitive to camera framing and the chosen off-screen-player policy.

\section{Limitations and Future Work}
\label{sec:limitations}

Identity fragmentation in real GSR output (track fragments; jersey-number OCR
coverage $\approx$45\% in our pipeline) makes roster-aware ghosting approximate
on real video: ghosts anchor to roles, not to verified identities. Appearance
re-identification is the natural next step. The pitch-control model is
velocity-free here to isolate the imputation effect; velocity-aware control with
imputed velocities is open --- the ablation shows short-gap extrapolation is
mildly informative for position, but its effect on velocity-aware control is
untested. The benchmark uses three matches from a single provider; more
matches and leagues would tighten the estimates, though the effect sizes are
large and consistent across all three matches and every viewport width. The simulated
viewport pans but does not zoom or tilt; zoom changes the number of visible
players, so real-broadcast occlusion statistics (including the 50--57\%
long-occlusion share) may differ from the simulated ones. All hyperparameters
($\tau = 8$\,s, EMA weight 0.1, the three-voter threshold, the 1.5\,s blend
constant) are round values fixed once during development on games 1--2 and not
optimised by search; the held-out game 3, evaluated only after they were
frozen, replicates the broad improvement over B0 and the effect magnitudes,
though the share-error ranking tightens (B2/B3V edge B4 by 0.3\pp{}). Finally,
learned imputers remain preferable when future context and training data are
available --- our contribution is the training-free floor, the harsher
online/long-occlusion benchmark, and the downstream-metric framing.

\section{Reproducibility}
\label{sec:repro}

All benchmark experiments run on CPU with open data (Metrica Sports sample
data~\citep{metrica2020}) and open code; the benchmark, the imputation ladder,
the figures, and the exact commands are available at
\url{https://github.com/nowayfootball/offscreen-impute}. The broadcast case
study uses World Cup footage that we cannot redistribute. The benchmark code is
publicly available, while the case-study pipeline and footage are not included
in the public repository. No GPU or cloud
expenditure was used for any experiment in this paper.

\bibliographystyle{plainnat}
\bibliography{refs}

\end{document}